\documentclass[12pt]{article}
\usepackage{amsmath,amssymb,amsfonts}
\usepackage{graphicx}
\usepackage{tabularx}
\usepackage{xcolor}
\usepackage{hyperref}
\usepackage{caption}
\usepackage{subcaption}
\usepackage{float}
\usepackage{mathrsfs}
\usepackage{cite}
\usepackage{url}
\usepackage{microtype}
\usepackage{booktabs}
\usepackage{authblk}
\usepackage{enumitem}

\usepackage[margin=1in]{geometry}

\title{A new framework for Marketing Mix Modeling: Addressing Channel Influence Bias and Cross-Channel Effects}

\author{Javier Marín}

\date{Original: January 15, 2023 \\ Revised: March 15, 2025}

\begin{document}

\maketitle
\begin{abstract}
This research addresses two fundamental challenges in Marketing Mix Modeling: the tendency of models to over-attribute influence to high-investment channels and the difficulty in quantifying cross-channel effects. We propose integrating the Michaelis-Menten equation and Maxwell-Boltzmann kinetic theory into hierarchical Bayesian models to overcome these limitations. Our approach uses the Michaelis-Menten model to characterize shape effects with spending-independent parameters and Boltzmann-type equations to systematically quantify cross-channel dynamics. Experimental results show that this physics-inspired approach maintains predictive accuracy while providing superior analytical insights into channel effectiveness and interactions. The normalized Michaelis-Menten constant offers an investment-independent measure of channel efficacy, while the N-particle system simulation reveals previously ignored channel interdependencies, enabling more accurate attribution and informed resource allocation decisions.

\vspace{0.5em}
\end{abstract}

\section{Introduction}

Marketing professionals use media mix models to assess the performance of their advertising campaigns. These models also provide insights to make future budget allocation decisions. Advertising influence typically has seasonality effects as well as other effects, like the shape effect, competitive effect, carryover, and lag \cite{Sethuraman2018, Tellis2004}. 

The shape effect is probably one of the most important insights for advertisers and refers to the change in sales in response to increasing intensity of advertising. This effect is commonly referred to as the saturation effect, and assumes that consumer response (in terms of sales, for example) tends to saturate when reaching a given point when increasing advertising effort. The most common approach to model this saturation curve argues that brands should expect a low responsiveness to low investment levels of advertising, a maximum responsiveness of sales at moderate levels, and a decreasing one at higher levels following a sigmoid shape. Predominant models fit sales response to S-shaped curves \cite{Tellis2004}.

Another important effect is the competitive effect, that measures the effect of a brand's advertising effectiveness relative to that of the other brands in the market. The carryover effect is the portion of advertising effectiveness that happens after the pulse of certain advertising. Last, the lag effect is the response delay to a given advertising input. These effects have been studied thoroughly over the last forty years, but the disruption of online media channels has utterly changed our understanding in how consumers interact with advertising. The digital world and social media have changed the way we interact with others, opening up a new understanding of advertising effects.

\section{Hill's equilibrium theory and shape effects}

Causal inference of advertising effects constitutes a critical dimension within marketing analytics, particularly for measuring incremental impacts and optimizing future campaign strategies. Models that fail to account for underlying consumer-advertising dynamics may introduce significant biases in causal estimates. Among these dynamics, the shape effect—characterizing the non-linear relationship between advertising intensity and sales response—warrants particular attention. Marketing practitioners widely acknowledge that consumer responsiveness follows a saturation pattern whereby, beyond certain threshold intensities, marginal returns diminish substantially as consumers become increasingly desensitized to additional message exposure. This saturation effect has critical implications for return on advertising investment planning, rendering a useful quantification tool for marketing resource allocation decisions. 

There are several theories in social psychology that could describe this phenomena, but the most commonly used approach to describe this effect is coming from a physics-inspired model: the Hill's equilibrium theory \cite{Gesztelyi2012}. This model was first used in 1909 to describe the interaction of hemoglobin with oxygen. Researchers observed that increasing the concentration of oxygen above a certain level could only slightly increase the amount of resulting interactions due to a saturation effect. Hill's equilibrium theory is comparable to Langmuir's theory of gas adsorption on solid surfaces, which also includes the saturation phenomenon \cite{Swenson2019}. The Hill equation can be derived from the law of mass action involving a ligand $L$, and a receptor $R$:

Reaction:
\begin{align}
nL + R \underset{k_2}{\overset{k_1}{\rightleftharpoons}} L_nR
\end{align}

Equations:
\begin{align}
[L_nR] &= [R_T] \cdot \frac{[L]^n}{[L]^n + K_d} \\
&= [R_T] \cdot \frac{[L]^n}{[L]^n + (K_A)^n}
\end{align}

where L is the ligand, $R$ is the receptor, $[L_nR]$ is the concentration of the ligand–receptor complex, $[R_T]$ is the total receptor concentration, $[L]$ is the concentration of the free ligand, $k_1$ and $k_2$ are reaction constants, $K_d = k_2 / k_1$ is the equilibrium dissociation constant of the ligand–receptor complex, $K_A$ is the ligand concentration at which half the receptors are occupied, and $n$ is the number of binding sites for the given ligand in one receptor, also called the Hill's coefficient or Hill's slope factor.

\begin{figure}[H]
\centering
\includegraphics[width=0.5\textwidth]{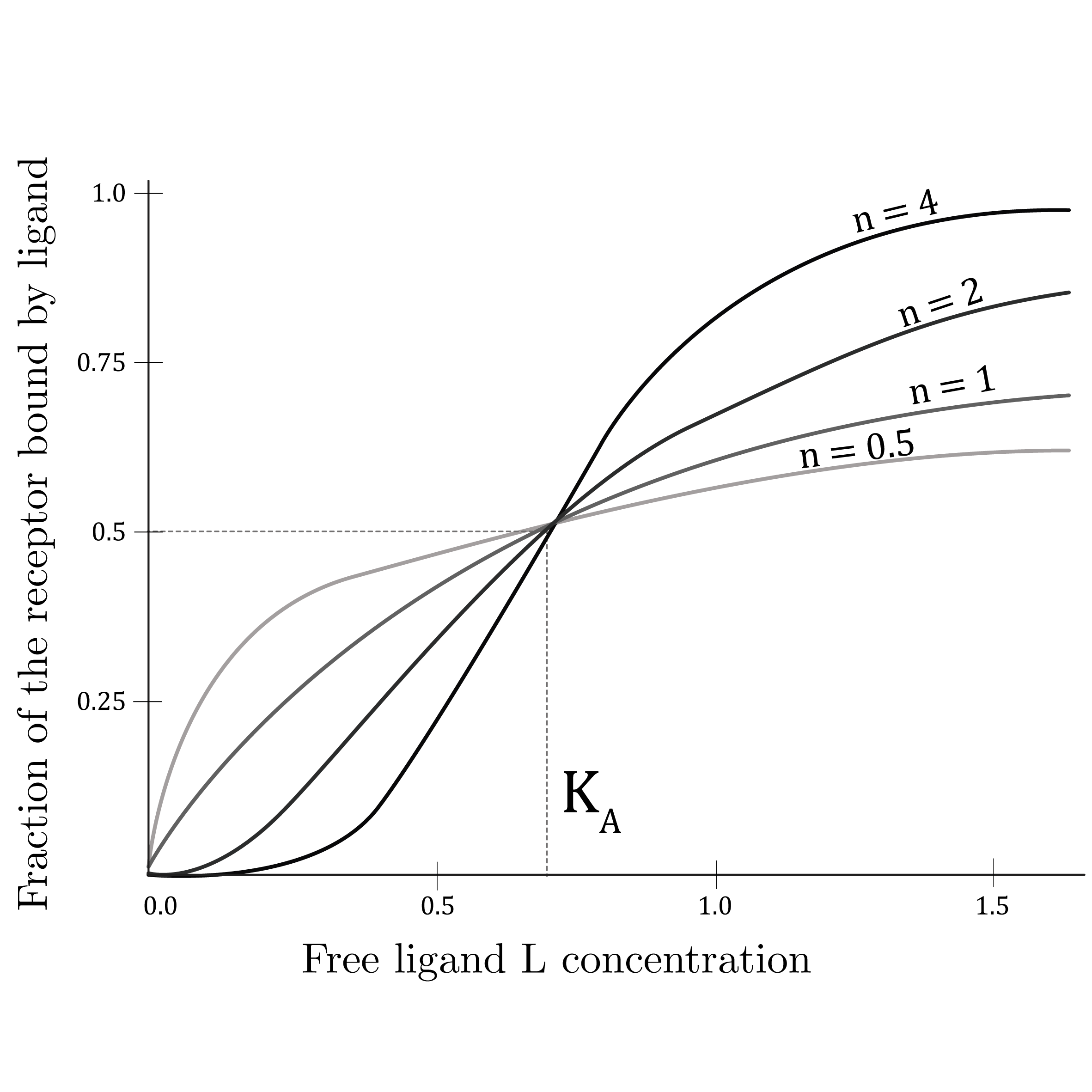}
\caption{Hill's model represented for different coefficients values $n$.}
\label{fig:hill-model}
\end{figure}

In figure \ref{fig:hill-model} we can see represented the different curves representing the fraction of receptor bound by ligands and the number of available ligands. As we can see, Hill coefficient $n$ strongly conditions the shape of this curve. For example, for high values, Hill equation shows a S-shape curve. Most advertisers consider the S-shape as the most plausible shape for advertising spend increase since it suggests that, at some very low level, advertising might be less effective since it gets blurred with noise. Researchers studying shape effect analysis have explored the advertising threshold as a potential rationale for the S-shaped curve. However,researchers have found the existence of a threshold in customer response to be highly improbable in reality. \cite{Bemmaor1984, Hanssens2003, Huang2012, Johansson1979, Simon1980, Tull1986, Vakratsas2004}. 

\section{Michaelis–Menten model and the advertising shape effect}

This model considers some important assumptions \cite{Gesztelyi2012} that can be translated to the specific application in advertising:

\begin{itemize}
\item The free ligand concentration is assumed not to differ significantly from the total ligand concentration. It's like assuming most of the ligands aren't already taken by receptors before a new ligand even start looking for a receptor. This simplification would means that all consumers has to react to ads spend at the same time, hypothesis that is not meaningful.

\item A ligand can only be bonded with a receptor if all available receptors are occupied. Imagine you're into gaming, and you've got this cool multiplayer game. Now, you know how sometimes you can team up with other players to get extra points or power-ups? Well, let's say there's a rule in the game that your team only gets a bonus if all your teammates activate their power-ups at the exact same time. Now, someone figured out that if not everyone activates their power-ups simultaneously, it messes up the whole bonus system. The equations they use to calculate these bonuses assume that everyone's powers are in sync. If some powers aren't fully activated, Hill's model can't give accurate results. For consumer response advertising modeling this condition also does not seem to be plausible.

\item Increasing affinity of the receptor after the binding of the first ligand accelerates the binding of the further ligands. This phenomena is known as homotropic cooperation. It would mean that consumers interact with each other necessarily in a way that new coming consumer are influenced by existing consumers that have already interacted with advertising spends. Assuming this may be reasonable, given that a customer can speak well about a certain brand, which may influence another customer's decision to choose or avoid that brand.
\end{itemize}

\section{Michaelis–Menten model and the advertising shape effect}

The Hill model is used to describe consumer responses to advertising since of the common idea of saturation or "reaching a certain limit". Saturation is a common feature in various physical phenomena and can be modeled mathematically. This concept is used to describe scientific phenomena such as how molecules adsorb onto a solid surface (Langmuir Adsorption Isotherm theory), the absorption of photons by electrons in a semiconductor, or the relaxation of nuclear spins in Nuclear Magnetic Resonance (NMR) among others. Despite the fact that the different "saturation" concepts are fundamentally different, and that it is not a direct or precise analogy, it has proved quite effective in explaining the consumer reaction curve or shape effect in marketing \cite{Tellis2004}. 

\setlength{\parskip}{12pt}

In this section, we will look at another important physics-inspired model commonly used in biology and chemistry that reproduces the same concept of saturation through a more complex approach. The Michaelis-Menten model explains enzyme catalysis's rate based on substrate concentration \cite{Lopez2000}. Catalysis by enzymes is a key mechanism in many biology and chemistry processes. Enzymes are specialized helpers that speed chemical reactions by generating an optimal environment for reactions and minimizing the amount of energy required for the reactions to take place. Both, ligand-receptor binding in Hill's model and substrate-enzyme interactions in Michaelis-Menten equation are selective, reversible, and saturable. It's important to note that Michaelis-Menten is equivalent to Hill's equation with coefficient $n = 1$ but with different parameters \cite{Gorban2011}. Originally, Michaelis and Menten only proposed that the substrate-enzyme reaction was in equilibrium. Briggs and Haldane, after adding the kinetic constants to chemical reactions, proposed a complete kinetic model \cite{Srinivasan2022}. Eventually, Stueckelberg extended that model to incorporate the Markov kinetics-described transitions of the intermediate compounds \cite{Gorban2011}. Following forward, we will look into these kinetic models in detail. 

The general Michaelis-Menten approach from Briggs and Haldane can be written as follows:

\begin{equation}
\begin{aligned}
\text{Chemical equilibrium:} \quad & S + E \underset{k_{-1}}{\overset{k_1}{\rightleftharpoons}} SE \xrightarrow{k_2} P + E \\
\\
\text{where:} \quad & [ES] = \frac{[E][S]}{K_M + [S]} \\
& \frac{d[S]}{dt} = -k_1[E][S] + k_{-1}[ES] \\
& \frac{d[ES]}{dt} = k_1[E][S] - (k_{-1} + k_2)[ES] \\
& \frac{d[P]}{dt} = k_2[ES] = \frac{k_2[ES]}{K_M + [S]}
\end{aligned}
\end{equation}

where $S$ is substrate, $E$ the enzyme or reactant, $ES$ is the enzyme-substrate ligand and $K_M$ is the Michaelis-Menten constant, providing a measure of the substrate concentration $[S]$ required for significant catalysis to take place. In equation 4, $[E] + [ES] = e = constant$ and $[S] + [P] = s = constant$. The condition $e \ll s$ is known as the Briggs-Haldane condition, implying a rapid equilibrium approximation \cite{Gorban2011}. According to Briggs and Haldane, constant $K_M$ can be written as

\begin{equation}
K_M = \frac{k_{-1} + k_2}{k_1}
\end{equation}

and is defined as the concentration of reactant $[E]$ at which the initial velocity of the enzymatic reaction is half of its initial velocity at saturation. Kinetic constant $k_1$ is the association rate constant of enzyme-substrate binding, and $k_{-1}$ is the rate constant of the $[ES]$ complex dissociating to regenerate free enzyme and substrate, and $k_2$ is the rate constant of the $[ES]$ complex dissociating to give free enzyme and product $[P]$. The equation of an enzymatic reaction (equation 4) is quite more complex than the Hill's model described in equation 3. Michaelis-Menten constant $K_M$ in equation 4 is only apparently equivalent with $K_A$, being $K_M$ also dependent on the rate constant of product formation in addition to the affinity of the substrate to the enzyme. 

The Briggs-Haldane extension to the Michaelis-Menten model incorporates several key assumptions: steady-state approximation, free ligand approximation, and rapid equilibrium approximation. The original Michaelis-Menten formulation posits that $[ES]$ approaches constancy when $[S] \gg K_M$ , which establishes the saturation phenomenon. Conversely, Briggs-Haldane introduced the alternative constraint $[E] + [ES] \ll [S]$. Both theoretical frameworks maintain that the reaction $[E] + [S] \rightleftharpoons [ES]$ achieves equilibrium, involving transient intermediate complexes $[ES]$ that simultaneously equilibrate with reactants $[ES]$ and $[S]$ while existing at negligible concentrations.

\begin{figure}[H]
\centering
\includegraphics[width=0.5\textwidth]{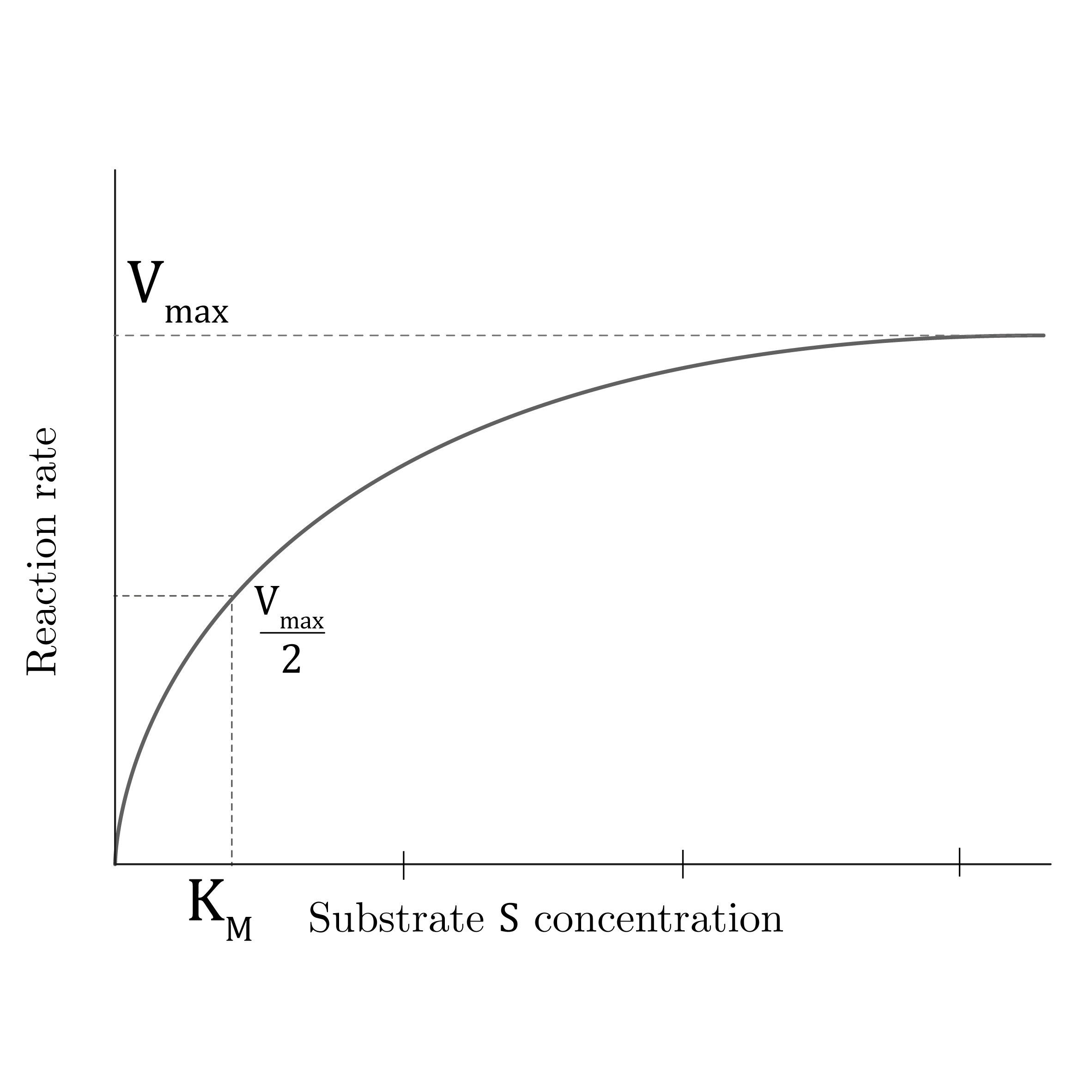}
\caption{Michaelis-Menten model curve representation.}
\label{fig:michaelis-menten}
\end{figure}

As we have noted, Michaelis and Menten first considered there were no interactions between intermediate complexes $[ES]$ due to their tiny amount and volume. Briggs and Haldane, however, posited that intermediate concentrations remain sufficiently elevated to facilitate significant interactions, maintaining their efficacy despite diminishing concentration gradients. As a result, nonlinear intermediate kinetics emerges, showing bifurcations, oscillations, and other complex behaviors \cite{Gorban2011}. In line with this, Stueckelberg made an important contribution to the understanding of intermediate collisions, particularly in the context of the Maxwell-Boltzmann equation \cite{Morgenstern1955}. Stueckelberg conceptualized quasi-chemical reactions as elastic collision events \cite{Gorban2011}, modeling intermediate complexes as binary particle systems with asymmetric velocities ($v$,$w$), whose post-collision kinetic properties shows probabilistic transformations \cite{Heer2012}. His formalism used linear Markov kinetics to characterize the temporal evolution of collision-derived compound states. In the Michaelis-Menten-Stueckelberg approach, transition kinetics are in fast equilibrium, intermediate concentrations are low, and transitions between them follow a linear kinetic equation \cite{Gorban2015}. In the Briggs and Haldane approach, transition kinetics are not in fast equilibrium.

\section{Michaelis-Menten Stueckelberg equation and behavioural sciences}

Approaches such as the Hill and Michaelis-Menten models could provide practical insights on media advertising saturation. As explained, these models are rooted on the premise that advertising messages reach a "saturation point", and while both models offer valuable perspectives, the Michaelis-Menten model stands out as a more comprehensive and detailed framework, specially Stueckelberg approach. This approach has two major advantages over the original Michaelis and Menten proposal: it is a comprehensive kinetic model that includes all of the kinetic constants, as well as intermediate collisions. These advantages will be practical when we establish and analogy with consumer behaviour to advertising. When we compare the mass balance equations produced from Hill's and Michaelis-Menten models, we can see that Hill's model could be analogous to a $stimulus \rightarrow response$ process. However, Michaelis-Menten model (both Briggs \& Haldane, and Stueckelberg) introduces a higher degree of complexity being able to capture and non-linear effects in modeling this saturation effect.

A general agreement is that in a socio-economically dynamic system, consumer intentions are predicted by attitudes and attitude change, personal and social norms, and perceived behavioural control \cite{Oinas-Kukkonen2013}.

\begin{figure}[H]
\centering
\includegraphics[width=0.5\textwidth]{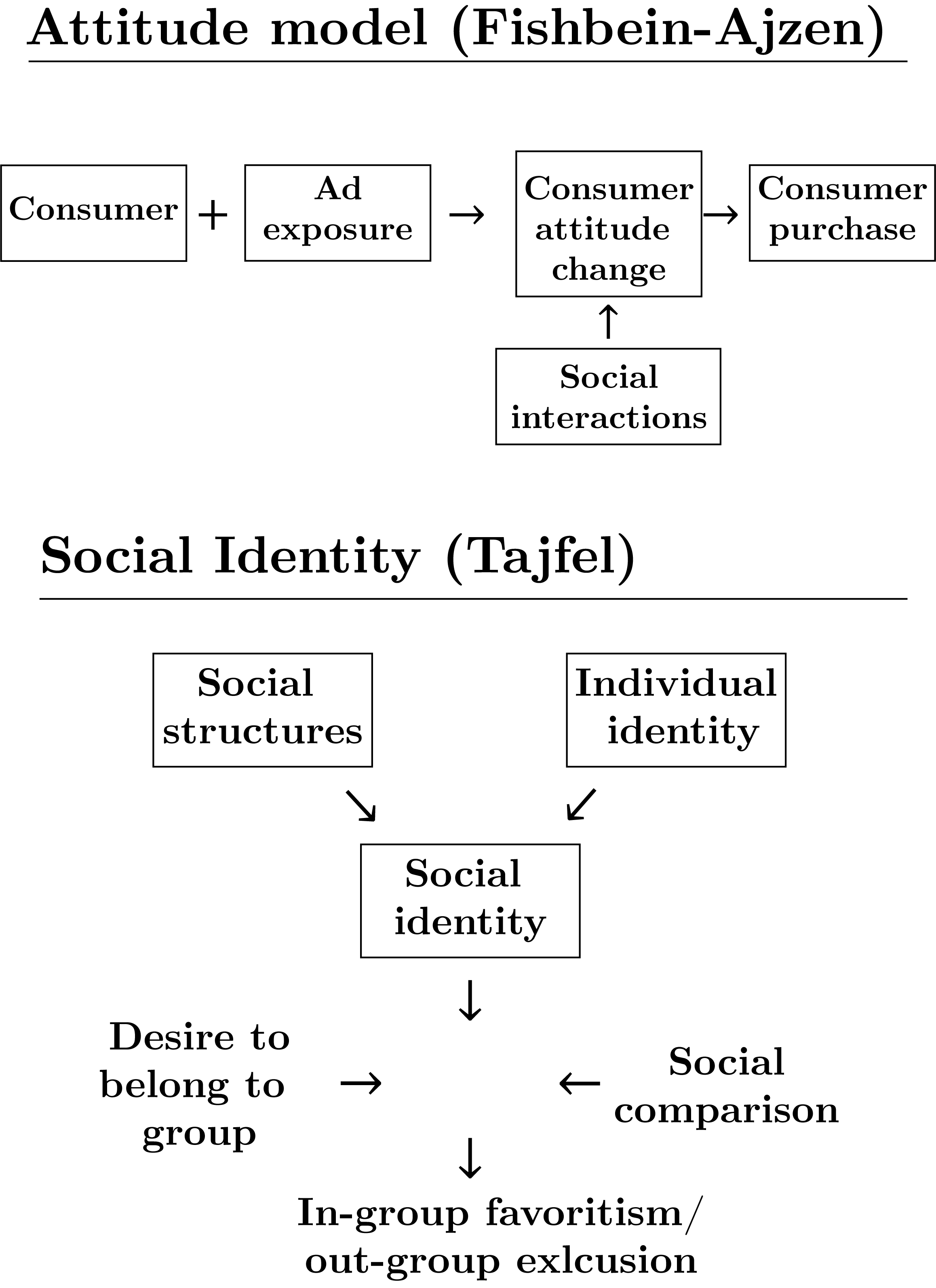}
\caption{Michaelis and Menten, Attitudes model and Social Identity model}
\label{fig:attitudes-model}
\end{figure}

Some recently developed conceptual work on consumer attitudes emphasizes the mediating role of consumer attitude between advertising and purchase intention \cite{Alalwan2018, Lewis2011, Xin2017}. This approach could be analogous to the mass action schema involving substrate $S$, and enzyme $E$ in figure \ref{fig:attitudes-model}. In the Michaelis-Menten equation version proposed by Briggs and Haldane, the constant $K_M$ can be understood in two different approaches:

\begin{itemize}
\item $K_M$ measures the strength of $[ES]$ complex: the lower the $K_M$ value, the tighter the substrate binding, or
\item $K_M$ measures the affinity of $[E]$ for $[S]$ when $k_{-1} \gg k_2$ in equation 12.
\end{itemize}

Using this analogy, the Michaelis-Menten constant $K_M$ in the consumer response curve represents a quantitative measure of consumer affinity for advertising in a specific channel. If $K_M$ grows, the consumer affinity for the channel decreases because larger spends are required to reach half of the maximum return value (the saturation point). Also $K_M$ might be seen as the "strength" of the advertising-consumer interaction through this channel. This "strength" might be compared to consumer attitude toward a specific product/brand, using the terminology used by Fishbein and Ajzen in their attitude model \cite{Ajzen2000}, or to the in-group favoritism described in the Tajfel's social identity model \cite{Hornsey2008} as shown in figure \ref{fig:attitudes-model}. In the Stueckelberg approach, the intermediate transitions of elementary reactions may be related to multiple stages influencing a potential consumer's path from being exposed to an ad to becoming actively engaged and finally converted. These stages are analogous to intermediate complexes (also called in the literature activated complexes). In this analogy we might have the equivalent in complex reactions with at least three "intermediates": exposed, engaged and converted.

\begin{align}
Ad_{Ch\ a} + Consumer_{Ch\ a} \rightleftharpoons Exposed_{Ch\ a} \rightleftharpoons Engaged_{Ch\ a} \rightleftharpoons Converted_{Ch\ a}
\end{align}

The intermediate transition states in Equation 4 have two main properties \cite{Gorban2021}:

\begin{itemize}
\item Quasi-steady state condition. The intermediate states have a brief lifetime and are present at much lower concentrations than the primary reagents. It would means that advertising and consumer interactions create intermediate states (exposed, engaged) that exist transiently compared to the persistent presence of advertising content and consumers. These short intermediate states rapidly transform either backward to their original states or forward toward conversion, without accumulating in significant quantities within the system.
\item Fast equilibrium approximation. The intermediate states establish rapid partial equilibrium with their preceding states, with reverse transitions occurring at rates substantially exceeding forward progression rates. This implies that consumers move fluidly between exposure and non-exposure, or between engagement and mere exposure, much more easily than they advance toward conversion—a phenomenon analogous to the thermodynamic preference for lower energy states in chemical systems.
\end{itemize}

These properties suggest rapid transitions between exposure, engagement, and conversion states. The Stueckelberg formalism enables modeling of both inter-consumer interactions among individuals exposed to identical advertising stimuli and the complex consumer-group behavioral dynamics through Boltzmann-type equations.

To support this analogy, we can consider that a consumer or social group's liking for a brand or product accelerates the transition from exposure to engagement. The intensity of an individual's affiliation with a social group influences this shift even more, especially when the brand is well aligned with the group's beliefs \cite{Farivar2022, Spears2011}. As individuals seek approval within their social circles, social validation works as a catalyst, driving the transition from passive exposure to active participation \cite{Brown2019}. Group identity dynamics influence how individuals allocate attention as they move from attracting attention to real conversion \cite{Chen2009}. Within-group social comparison, together with in-group affinity, plays a key role for evaluating the possibility of attention turning to conversion \cite{Wachter2020}. When transitioning from generating interest to making a purchasing choice, social proof becomes fundamental a driving force.

\section{A hierarchical Bayesian framework for Marketing Mix Modeling}

Bayesian models have shown a robust and flexible approach to analyze the change in sales caused by an exposure to advertising \cite{Chen2021, Fahrmeir2001, Jin2017, Singmann2019, Wang2017}. One of the principal methodological advantages of Bayesian frameworks lies in their capacity to formally incorporate domain expertise through informative prior distributions on parameters, thereby integrating established theoretical knowledge and empirical findings from previous research into the estimation process. This prior specification mechanism enables the systematization of accumulated scientific understanding, facilitates more efficient parameter estimation in high-dimensional spaces, provides natural regularization against overfitting, and allows for principled uncertainty quantification—advantages that are particularly salient when modeling complex marketing phenomena with inherently limited observational data. 

A commonly used Bayesian model for Marketing Mix Modeling was introduced by Jin et al. from Google Research in 2017\footnote{This library is available in https://github.com/google/lightweight\_mmm} \cite{Jin2017}. This model uses a Bayesian inferential framework with Markov Chain Monte Carlo (MCMC) simulation methods \cite{Cheung2009}. The framework has domain-specific prior knowledge about different advertising effects, including carryover and shape effects, while simultaneously accounting for baseline effects through an intercept parameter, temporal cyclical via seasonality components, and longitudinal evolution through global trend parameters. The model architecture accepts multidimensional input data including a temporal index column for chronological ordering, multiple advertising expenditure variables across diverse channels, additional control variables, and a dependent response variable that quantifies marketing performance. The dependent variables can be represented as:
\begin{equation}
X = (x_1, x_2, ..., x_m)
\end{equation}
and the control variables as:
\begin{equation}
Z = (z_1, z_2, ..., z_n)
\end{equation}

These control variables may be any measure that can influence the dependent variable $y$. Some examples of control variables could be macro-economic variables, weather data, and competitors information. This data can be daily or weekly data and national or geo-level data. The general equation describing this model is the following:

\begin{equation}
p(\Phi|y, X) \propto L(y|X, Z, \Phi)p(\Phi)
\end{equation}

where $y$ is the dependent variable, $X = (x_1, x_2, ..., x_m)$ and $Z = (z_1, z_2, ..., z_n)$ are the media and control variables (often referred to as non-media variables), $\Phi$ the model parameters. The terms $p(\Phi|y, X)$ is the conditional distribution of observed data $y$, and $L(y|X, Z, \Phi)$ is the log likelihood given the data and the parameters. The simplified model formulation can be represented re-written as \cite{Jin2017}:

\begin{equation}
y(t) = \alpha + trend(t) + seasonality(t) + media(t) + non-media(t)
\end{equation}

where $\alpha$ is a constant or intercept and the outcome is the dependent variable $y$. The most important component in this equation is the media spend, where we would like to introduce information about some effects like carryover and shape effect.

\subsection{Advertising funnel effects and the Boltzmann kinetic theory}

Funnel effects in advertising refer to the cross-effects between advertising channels, where one channel's activity influences the effectiveness of another. When these interactions are not properly accounted for, marketing mix models produce biased influence estimations \cite{Chan2017}. Despite broad consensus that channels impact one another, particularly in digital environments \cite{Gordon2020}, this idea lacks explanatory models that effectively quantify these interactions. This section introduces a conceptual framework derived from non-linear Boltzmann-type equations \cite{Morgenstern1955} to evaluate cross-channel interactions in advertising. The application of these equations is justified by their proven capacity to characterize dynamic and stochastic interactions in complex systems. We propose to apply it to socio-economic systems, providing a rigorous framework for understanding advertising cross-channel dynamics \cite{Toscani2019}.

The Maxwell-Boltzmann kinetic theory of gases\cite{Morgenstern1955} provides a basic formalism for particle behavior through statistical mechanics. In this formalism, two-particle interactions follow classical conservation laws of momentum and kinetic energy:

\begin{align}
v_1 + v_2 &= v_1^* + v_2^* \\
|v_1|^2 + |v_2|^2 &= |v_1^*|^2 + |v_2^*|^2
\end{align}

where $v$ represents particle velocity for particles $1$ and $2$. The parametrized solution to this system of algebraic equations is given by \cite{Dimarco2014}:

\begin{align}
v_1^* &= \frac{1}{2}(v_1 + v_2 + |v_1 - v_2|\omega) \\
v_2^* &= \frac{1}{2}(v_1 + v_2 - |v_1 - v_2|\omega)
\end{align}
where $v_1 - v_2$ is the relative velocity and $\omega$ is a unit vector of the sphere $S^2$ (for two channels). This theoretical framework's postulation that gas molecules exist in constant, random motion with frequent collisions provides an apt metaphor for consumer behavior in marketing contexts. Consumers continually encounter advertising from multiple sources, with their responses influenced by complex interactions among these stimuli.

Considering the distribution function $f = f(v,t)$ for $t > 0$, and applying the molecular chaos hypothesis \cite{Morgenstern1955}, the evolution of $f$ (and observable quantity $\varphi(v)$) can be described by a highly non-linear Boltzmann-type equation \cite{Toscani2019}:

\begin{equation}
\frac{d}{dt}\int_{\mathbb{R}} \varphi(v)f(v,t) dv = \frac{1}{\tau N} \int_{\mathbb{R}^N}\sum_{i=1}^{N}\langle\varphi(v_i^*) - \varphi(v_i)\rangle_{\tau}\prod_{j=1}^{N}f(v_j,t)dv_1...dv_N
\end{equation}
where $i,j$ denotes particles and $\tau$ is the relaxation time. This equation can be generalized using numerical methods \cite{Dimarco2014, Toscani2019}. For $N$-particles ($N \gg 1$), with pre-collisional state $V = (v_1, v_2, ..., v_N)$ and post-collisional state $V^* = (v_1^*, v_2^*, ..., v_N^*)$, we can describe $V^*$ as a random linear transformation of $V$ \cite{Toscani2019}:

\begin{equation}
v_i^* = av_i + b\sum_{j=1}^{N}v_j
\end{equation}
for $i = 1, ..., N$, where parameters $a$ and $b$ can be fixed or bounded randomly distributed. 

In the advertising context, channels are conceptualized as $N$ particles, with $v$ denoting their influence on dependent variable $y$. The vector $V$ represents initial channel influences (measured in isolation), while post-collisional state $V^*$ captures adjusted influence values incorporating cross-channel interactions. The adaptation of Equation 15 requires replacing $\mathbb{R}$ with $\mathbb{R}^+$ if we consider the assumption that channel influence is non-negative. Both equations 15 and 16 are aligned with Maxwell's molecular chaos hypothesis \cite{Maxwell1986}, when $N \gg 1$.

A hierarchical Bayesian model generates an output matrix that can be represented as:
\begin{equation}
\begin{pmatrix}
v_{1,t=1} & \cdots & v_{N,t=1} \\
\vdots & \ddots & \vdots \\
v_{1,t=T} & \cdots & v_{N,t=T}
\end{pmatrix}
\end{equation}
where $N$ is the number of advertising channels, and $t$ is the time step. For a model with the form $av_{1,t} + bv_{2,t} = z_t$, we can reformulate as $z_t = av_{1,t} + bv_{2,t} + \epsilon_t$, where $\epsilon_t$ represents the error associated with each observation. The likelihood function for a single observation follows a normal distribution:

\begin{equation}
f(\epsilon_t) = \frac{1}{\sqrt{2\pi\sigma^2}}\exp\left(-\frac{\epsilon_t^2}{2\sigma^2}\right)
\end{equation}

Assuming independent and identically distributed errors, the likelihood function for all observations becomes:

\begin{equation}
\mathcal{L}(M_o|a, b, \sigma^2) = \prod_{t=1}^{n}\frac{1}{\sqrt{2\pi\sigma^2}}\exp\left(-\frac{\epsilon_t^2}{2\sigma^2}\right)
\end{equation}
where $M_o$ represents observed data and $n = t$ is the number of time steps. Taking the logarithm, we get the expression:

\begin{equation}
\log \mathcal{L}(M_o|a, b, \sigma^2) = -\frac{n}{2}\log(2\pi\sigma^2) - \frac{1}{2\sigma^2}\sum_{t=1}^{n}\epsilon_t^2
\end{equation}
Parameters $a$, $b$, and $\sigma^2$ can be estimated by maximizing Equation 19 or minimizing the sum of squared errors. The new channel influence $v_i^*$ in Equation 16 can be estimated through sequential least squares programming \cite{Fu2019}, or an integrative approach incorporating Equation 16 as priors in a Bayesian hierarchical model.  The advantage of separate parameter calculation is avoiding increased parameter dimensionality in the Bayesian model, particularly important when we have multiple media channels \cite{Cheung2009, Spiegelhalter1998} and control variables are considered. This approach also enables paired cross-relationship analysis rather than aggregate influence assessment, though at the cost of implementation complexity.

\subsection{Integration of Michaelis-Menten and Boltzmann Equations in Media Mix Modeling}

 In this paper, we introduce a different approach in which we model the shape effect with the basic Michaelis-Menten equation, adding also funnel effects modeled using a Boltzmann-type equation. 
 
 The adstock model can be written as\cite{Jin2017}:

\begin{align}
adstock(x_{t-L+1,m}, ..., x_{t,m}; w_m, L) &= \frac{\sum_{l=0}^{L}w_m(l)x_{t-l,m}}{\sum_{l=0}^{L}w_m(l)} \\
carryover(x_{t-L+1,m}, ..., x_{t,m}; w_m, L) &= \frac{\sum_{l=0}^{L}\tau_m^{(l-\delta_m)^2}x_{t-l,m}}{\sum_{l=0}^{L}\tau_m^{(l-\delta_m)^2}}
\end{align}
where $L$ is the maximum delay time (Lightweight library documentation suggests 13 weeks), and $w$ is the delay adstock function,

\begin{equation}
w_m^a(l; \alpha_m, \delta_m) = \alpha_m^{(l-\delta_m)^2}
\end{equation}
for $l$ in $[0, L-1]$, $0 < \alpha_m < 1$ and $0 \leq \delta_m \leq 1$. In the carryover approach, $\tau$ is the ad effect retention rate. The funnel effect can be written in the following functional form:

\begin{equation}
Boltzmann(x_{i,m}^*; a_m, b_m) = a_mx_{i,m}^* + b_m\sum_{j=1}^{N-1}x_{j,m}^*
\end{equation}
where $x_{i,m}^*$ is the output from the transformation for a given channel with equations 21 or 22, and $\sum_{j=1}^{N-1}x_{j,m}^*$ is the sum of the transformation of the remaining channels $j$ at the same time step $t$. 

As explained in section 6.2, funnel effect model using Boltzmann-type equation can be solved using two different strategies: considering two-particle or $N$-particle interactions (low-order or high-order interactions). Equation 24 uses high-order interactions, meaning that we estimate channel influence change from the addition of interactions from all other channels at the same time. Note that this approach does not consider all particles colliding simultaneously like in original Maxwell-Boltzmann's kinetic theory. Collisions happen between two particles, not between multiple or all particles at the same time. This approach will not provide insights about what channel is influencing another. But marketers need to understand the media channel that is driving another media channel's influence change. In that case, we can use the approach provided by equation 24 but comparing channels $i$ and $j$ to calculate the influence change of $i$ as $v_i^* - v_i$ (post-collisional state minus original state):

\begin{equation}
v_i^* = av_i + bv_j
\end{equation}

Equation 25 allows quantification of channel-specific cross-effects when applied to hierarchical Bayesian model estimations that incorporate only carryover and shape priors. The shape effect, characterized through the Michaelis-Menten formalism, takes the form:

\begin{equation}
MichaelisMenten(x_{i,m}^{**}; V_m, K_m) = \frac{V_m x_{i,m}^{**}}{x_{i,m}^{**} + K_m} + \epsilon
\end{equation}
where $x_{i,m}^{**}$ is the output from equation 15 (can be from both approaches, adstock and carryover), $V_m$ is the maximum saturation, $K_m$ is the Michaelis-Menten constant at this value of concentration (when half of the maximum saturation is reached), and $\epsilon$ is the error term. Last, we can use this transformation output as input for the Michaelis-Menten model. 

Let $y_t$ be the response variable at time $t$, the final response can be modeled by the following generic equation:

\begin{equation}
y_t = \tau + \sum_{m=1}^{M}MichaelisMentenBoltzmann(x_{i,m}^{**}; V_m, K_m) + b\sum_{c=1}^{C}\gamma_cz_{t,c} + \epsilon_t
\end{equation}
where $\tau$ is the baseline, $\gamma_c$ is the control variable effect $z_{t,c}$, and $\epsilon_t$ is some noise that is assumed to be uncorrelated with the other variables in the model and to have constant variance.

As previously suggested, removing the funnel effect from the hierarchical Bayesian model will be a valid strategy for identifying paired funnel effects:

\begin{equation}
y_t = \tau + \sum_{m=1}^{M}MichaelisMenten(x_{i,m}^*; V_m, K_m) + b\sum_{c=1}^{C}\gamma_cz_{t,c} + \epsilon_t
\end{equation}

After training the model with only the carryover and shape effects, we get an output with the form:

\begin{equation}
\begin{pmatrix}
v_{1,t=1} & \cdots & v_{N,t=1} \\
\vdots & \ddots & \vdots \\
v_{1,t=T} & \cdots & v_{N,t=T}
\end{pmatrix}
\end{equation}
where $v_i$, for $1 < i < N$, represents channel influence, $N$ denotes the number of advertising channels, and $t$ indicates the time step. Equation 25 allows us to determine how channel $j$ impacts target channel $i$, measured as the influence change $v_i^* - v_i$, with $v_i^*$ representing the post-collisional state. This equation can be solved using sequential least squares programming.

\section{Validation with data}

We evaluated the hierarchical model as described in Equation 10 using two synthetic datasets: dataset 1, comprising ten media channels (online and offline), two control variables, and a dependent variable spanning 2.5 years of weekly observations for an online retailer with negative ROAS characteristics $y_t \ll x_{i,t}^*$; and dataset 2, representing a physical retail business with one online and five offline advertising channels. 

Dataset 1 can be represented as:

\begin{equation}
\begin{pmatrix}
t_1 & x_{1,t=1} & \cdots & x_{10,t=1} & z_{1,t=1} & z_{2,t=1} \\
\vdots & \vdots & \ddots & \vdots & \vdots & \vdots \\
t_{144} & x_{N,t=144} & \cdots & x_{10,t=144} & z_{1,t=144} & z_{2,t=144}
\end{pmatrix} \propto 
\begin{pmatrix}
y_{t=1} \\
\vdots \\
y_{t=144}
\end{pmatrix}
\end{equation}
where $t$ is the time step in weeks, $x_1...x_{10}$ is the weekly spend per channel, $z_1$ and $z_2$ are the control variables, and $y$ is the dependent variable.

\begin{figure}[H]
\centering
\includegraphics[width=0.5\textwidth]{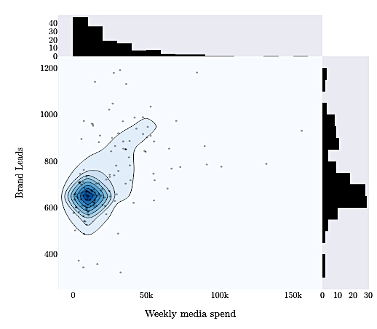}
\caption{Histogram contour plot comparing total weekly media spend and Brand leads (dataset 1)}
\label{fig:contour-plot}
\end{figure}

Our input data variables show very low correlation ($r^2 < 0.4$), avoiding for specific transformations to address potential limitations in the convergence of the MCMC sampler. Using a modified version of the original Lightweight library we compute two chains of the NUTS MCMC algorithm. Table \ref{tab:prediction-accuracy} presents the predictive results obtained from six distinct models: adstock, carryover, and Hill adstock (original models from the Lightweight MMM library), along with three Michaelis-Menten models described in equations 28 and 29. These Michaelis-Menten models include Michaelis-Menten + adstock, Michaelis-Menten + carryover, and Michaelis-Menten + Boltzmann. Michaelis-Menten-adstock incorporates the adstock model described in Equation 21 as input for Equation 24, while Michaelis-Menten-carryover uses the carryover model in Equation 22 as input for Equation 24. Throughout these experiments, consistent prior distributions have been applied to parameter values, ensuring comparability across all models.

\begin{table}[h]
\centering
\setlength{\tabcolsep}{4pt}
\footnotesize
\caption{Prediction accuracy of different models for dataset 1}
\label{tab:prediction-accuracy}
\begin{tabular}{lcccccc}
\toprule
Experiment & Adstock & Carryover & Hill-Adstock & MM+adstock & MM+carryover & MM+Boltzmann \\
\midrule
$R^2$ & 0.79 & 0.78 & 0.84 & 0.84 & 0.78 & 0.77 \\
Explained Variance & 0.79 & 0.78 & 0.84 & 0.84 & 0.78 & 0.77 \\
MAPE & 0.087 & 0.088 & 0.077 & 0.078 & 0.089 & 0.09 \\
Accuracy \% & 92.4 & 92.2 & 93.4 & 93.3 & 92.0 & 91.9 \\
\bottomrule
\end{tabular}
\end{table}

In Table \ref{tab:prediction-accuracy} we can see the results of the original $y$ forecast using different models. We can observe that models using the adstock equation to quantify the channels' influence (Hill-Adstock and Michaelis-Menten+adstock ) have slightly higher accuracy than models that use the carryover effect (Carryover and Michaelis-Menten+carryover). 

\begin{table}[h]
\centering
\setlength{\tabcolsep}{4pt}
\footnotesize
\caption{Media channel's contribution (in \%) for the different experiments with dataset 1}
\label{tab:media-contribution}
\begin{tabular}{lccccc}
\toprule
 & Carryover & Hill-Adstock & MM+carryover & MM+adstock & MM+Boltzmann+carryover \\
\midrule
Ch1 offline & 2.46 & 5.6 & 1.50 & 0.37 & 0.58 \\
Ch2 offline & 1.92 & 2.64 & 0.90 & 0.90 & 0.53 \\
Ch3 online & 0.65 & 5.7 & 0.80 & 0.64 & 0.61 \\
Ch4 offline & 0.69 & 2.49 & 0.40 & 0.90 & 0.48 \\
Ch5 online & 0.59 & 7.1 & 0.50 & 0.43 & 0.68 \\
Ch6 online & 5.33 & 7.41 & 3.60 & 0.89 & 2.33 \\
Ch7 online & 1.73 & 11.18 & 1.00 & 1.11 & 0.82 \\
Ch8 online & 0.09 & 0.39 & 0.10 & 0.00 & 0.28 \\
Ch9 offline & 0.26 & 38.94 & 0.10 & 45.47 & 0.27 \\
Ch10 online & 0.14 & 0.64 & 0.10 & 0.00 & 0.29 \\
Total & 13.86 & 82.09 & 9.00 & 50.71 & 6.87 \\
\bottomrule
\end{tabular}
\end{table}

Table \ref{tab:media-contribution} shows the results of the estimation channel's influence. Notably, results obtained from the original Lightweight carryover, Michaelis-Menten+carryover, and Michaelis-Menten+Boltzmann models shows clear similarities, with total media contributions of 13.86\%, 9.0\%, and 6.87\%, respectively. However, these results markedly differ from those generated by the model based on the adstock channel's influence equation, showing contributions of 82.9\% for the Lightweight MMM library model and 50.71\% for Michaelis-Menten+Adstock. This variance is expected and means that the adstock equation interprets channel influence in a distinct manner. Adstock approach deviates from carryover models by assuming that the maximum influence of each channel matches with the campaign's launch. This behaviour has been disagreeable among marketing analysts due to its divergence from observed results and experiential insights. The two model groups (with carryover and with adstock) shows clear different outcomes. Within the carryover model group, differences, albeit more moderate, emerge among the Lightweight original model, the Michaelis-Menten model with or without the addition of the Boltzmann formalism. In a broader context, our findings reveal that augmenting the number of parameters to estimate in the Bayesian model diminishes the estimated value of channel influence.

\begin{table}[h]
\centering
\setlength{\tabcolsep}{4pt}
\footnotesize
\caption{Michaelis-Menten $V_{max}$ parameter for the different experiments with dataset 1}
\label{tab:vmax}
\begin{tabular}{lccc}
\toprule
channel  & Michaelis-Menten+adstock & Michaelis-Menten+carryover & Michaelis-Menten+Boltzmann \\
\midrule
Ch1 offline & 41.87 & 38.83 & 31.07 \\
Ch2 offline & 39.24 & 32.87 & 31.37 \\
Ch3 online & 33.14 & 24.84 & 25.00 \\
Ch4 offline & 32.93 & 24.14 & 25.60 \\
Ch5 online & 34.19 & 24.73 & 26.97 \\
Ch6 online & 36.32 & 40.67 & 38.14 \\
Ch7 online & 37.18 & 30.79 & 30.20 \\
Ch8 online & 36.15 & 18.26 & 23.75 \\
Ch9 offline & 63.53 & 20.44 & 22.39 \\
Ch10 online & 36.68 & 25.87 & 26.13 \\
\bottomrule
\end{tabular}
\end{table}

\begin{table}[h]
\centering
\setlength{\tabcolsep}{4pt}
\footnotesize
\caption{Michaelis-Menten constant $K_M$ for the different experiments with dataset 1}
\label{tab:km}
\begin{tabular}{lccc}
\toprule
Media channel & Michaelis-Menten-adstock & Michaelis-Menten-carryover & Michaelis-Menten-Boltzmann \\
\midrule
Ch1 offline & 3,484 & 4,267 & 4,384 \\
Ch2 offline & 16,013 & 17,632 & 17,084 \\
Ch3 online & 13,314 & 13,666 & 13,275 \\
Ch4 offline & 23,700 & 26,635 & 25,527 \\
Ch5 online & 9,267 & 9,606 & 9195 \\
Ch6 online & 12,692 & 10,179 & 11,085 \\
Ch7 online & 17,802 & 19800 & 19,033 \\
Ch8 online & 111 & 119 & 116 \\
Ch9 offline & 12 & 35,237 & 32,870 \\
Ch10 online & 27 & 28 & 29 \\
\bottomrule
\end{tabular}
\end{table}

\begin{figure}[H]
\centering
\includegraphics[width=0.5\textwidth]{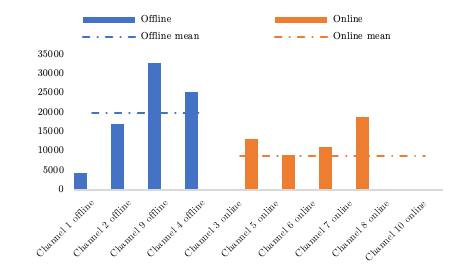}
\caption{Michaelis-Menten constant $K_M$ comparison for online and offline channels for dataset 1}
\label{fig:km-comparison}
\end{figure}

Model's outputs for the Michaelis-Menten equation parameters $K_M$ and $V_{max}$ are shown in tables \ref{tab:vmax} and \ref{tab:km}. Offline channels seems to have have higher saturation levels (figure \ref{fig:km-comparison}), but online channels have higher $y$ (RoAS -Return on Advertising Spend-) corresponding with higher $V_{max}$ values. As established in section 4, the Michaelis-Menten parameter $K_M$ go beyond its computational role in response curve approximation by offering interpretable connections to underlying consumer behavior mechanisms.

\begin{figure}[H]
\centering
\includegraphics[width=0.5\textwidth]{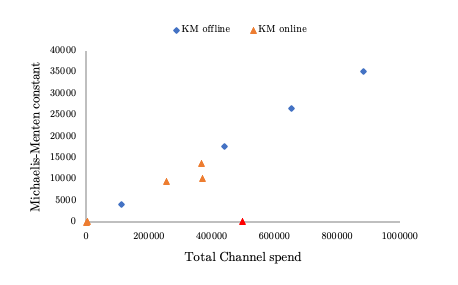}
\caption{Michaelis-Menten constant $K_M$ versus Total Channel Spend for online and offline channels for dataset 1}
\label{fig:km-vs-spend}
\end{figure}

As previously discussed, $K_M$ values reflect the affinity of consumers with a product/brand through a given channel, but with an inverse relation. However, to understand the true implications of $K_M$ values in the Michaelis-Menten equation, we must consider its unique behavior, identifying three different regions:

\begin{itemize}
\item Region 1:When $x \ll K_M$, the equation simplifies to $y \propto x$, indicating that outcomes are proportional to channel spend. Then $K_M$ is proportional to $x$ and to $\sum x$.
\end{itemize}

\begin{itemize}
\item Region 2: When $x \approx K_M$, then $y \approx V_{max}/2$, and $K_M$ remains proportional to $x$ and to $\sum x$.
\end{itemize}

\begin{itemize}
\item Region 3: For $x \gg K_M$, the full Michaelis-Menten equation applies: $y = V_{max}x/(K_M + x)$, being $K_M = x(V_{max}/y - 1)$
\end{itemize}

Given that our experiment's weekly spend values are predominantly lower than $K_M$, with some exceptions, our data is located within regions 1 and 2. To address the original Michaelis-Menten approach's limitation, $K_M$ is normalized by dividing each channel constant by the sum of weekly channel spend $(S_x)$.

The normalized constant $\tilde{K_M}$ provides a standardized measure of consumer-product affinity and exposure-to-purchase conversion rates. This normalization enables valid cross-experiment comparisons despite varying concentration levels, analogous to quantifying effective substrate contribution in enzyme-substrate complex formation. Consequently, $\tilde{K_M}$ serves as a robust metric for assessing relative channel importance.

\begin{figure}[H]
\centering
\includegraphics[width=0.5\textwidth]{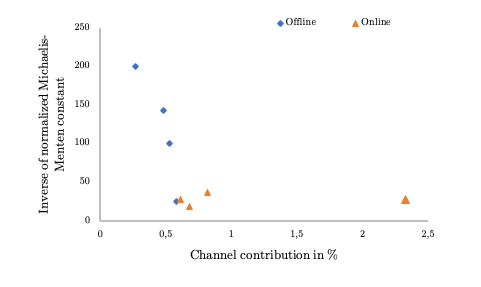}
\caption{Correlation between $\tilde{K_M}$ and channel influence for dataset 1}
\label{fig:km-correlation}
\end{figure}

Figure \ref{fig:km-correlation} shows the correlation between normalized constant $\tilde{K_M}$ and the channel influence obtained from the model. We can observe that there is not a direct correlation between the two values. Mostly, online channels presents higher normalized $\tilde{K_M}$, regardless of their influence, indicating lower affinity or reduced interaction-to-conversion rates. This implies that exposure to purchase decision rates is lower for online channels. The evidence suggests that the normalized Michaelis-Menten constant $\tilde{K_M}$ is less contingent on the overall spending in a specific channel. It is important to note that no positive correlation has been identified between a channel's RoAS and normalized constant $\tilde{K_M}$. Consequently, this constant or its inverse, is not valid to measure channel profitability.

\begin{table}[h]
\centering
\setlength{\tabcolsep}{4pt}
\footnotesize
\caption{Media contribution for dataset 2 calculated with different models}
\label{tab:media-contribution-ds2}
\begin{tabular}{lccccc}
\toprule
 & 'Carryover' & 'Adstock' & 'Hill Adstock' & MM+carryover & MM+Boltzmann+carryover \\
\midrule
Online 1 & 5.53 & 5.13 & 31.83 & 22.57 & 16.88 \\
Offline 1 & 3.7 & 3.47 & 3.5 & 2.96 & 4.21 \\
Offline 2 & 1.15 & 0.17 & 0.63 & 0.19 & 2.72 \\
Offline 3 & 10.43 & 15.14 & 9.34 & 7.82 & 6.08 \\
Offline 4 & 1.73 & 0.21 & 3.19 & 0.28 & 2.73 \\
Offline 5 & 28.33 & 28.94 & 31.24 & 43.14 & 40.54 \\
Total  & 50.9 & 53.1 & 79.7 & 77.0 & 73.2 \\
\bottomrule
\end{tabular}
\end{table}

Table \ref{tab:media-contribution-ds2} shows the results of the experiments with dataset 2 using standard Lightweight MMM library models and the experimental models described in the previous section. In contrast to what happens with dataset 1, the results of total channel influence do not differ significantly. At this point, an important consideration comes up: do the constant $\tilde{K_M}$ (normalized affinity, or exposure to purchase decision rate) and the channel contribution have different meanings? If there are actually distinct parameters, what are the Michaelis-Menten coefficients really suggesting? To answer these questions, we are going to use another dataset. We have modeled dataset 2 using Equation 14 described in the previous section. Results can be seen in table \ref{tab:experiment-results-ds2}.

\begin{table}[h]
\centering
\caption{Experiment results with dataset 2 and Michaelis-Menten model}
\label{tab:experiment-results-ds2}
\begin{tabular}{lccccc}
\toprule
Channels & Media spend & Contribution \% & Media outcome & CpA & $K_M$ \\
\midrule
Online 1 & 58921 & 22.57 & 13531 & 4.4 & 406 \\
Offline 1 & 49101 & 2.96 & 1333 & 36.8 & 337 \\
Offline 2 & 1348 & 0.19 & 3 & 499.4 & 6 \\
Offline 3 & 464983 & 7.82 & 37455 & 12.4 & 3014 \\
Offline 4 & 2919 & 0.28 & 8 & 351.6 & 14 \\
Offline 5 & 55527 & 43.14 & 24243 & 2.3 & 618 \\
\bottomrule
\end{tabular}
\end{table}

\begin{figure}[H]
\centering
\includegraphics[width=0.5\textwidth]{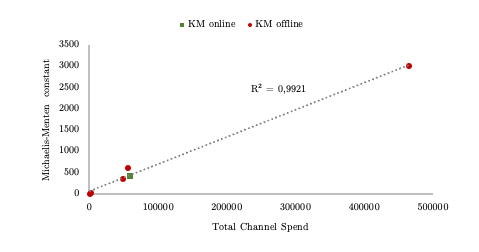}
\caption{Michaelis-Menten constant $K_M$ versus Total Channel Spend for online and offline channels for dataset 2}
\label{fig:km-vs-spend-ds2}
\end{figure}

As we have seen in the example with dataset 1, Michaelis-Menten constant is also correlated with the total spend per channel (figure \ref{fig:km-vs-spend-ds2}).

\begin{figure}[h]
\centering
\includegraphics[width=0.75\textwidth]{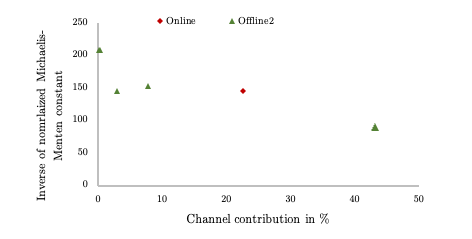}
\caption{Correlation between normalized Michaelis-Menten constant $\tilde{K_M}$ and channel influence for dataset 2}
\label{fig:km-correlation-ds2}
\end{figure}

Figures \ref{fig:km-correlation} and \ref{fig:km-correlation-ds2} indicate a lack of correlation between $\tilde{K_M}^{-1}$ with channel influence. This observation suggests a distinction between channel influence and the Michaelis-Menten constant (and its normalized inverse), indicating the potential representation of different phenomena. Noticeably, $\tilde{K_M}$ and $\tilde{K_M}^{-1}$ offer an advantage over media contribution values by remaining unaffected by the total spending on each channel. Our model, as represented by equation 10, establishes a non-direct relationship between media spend and Key Performance Indicator (KPI). In scenarios where a new channel is tested with a modest budget, the model's output predominantly reflects a low influence value for the channel. In contrast, normalizing the Michaelis-Menten constant provides a more accurate measure of the channel's conversion potential or affinity (for using the same terminology introduced by Michaelis and Menten) defined as $1/\tilde{K_M}$. Particularly, this factor differs from channel influence as it offers a nuanced measure that remains independent of total channel spend, thereby providing valuable insights across different spending levels. This is a very common limitation in MMM outcomes where we have extreme investment levels. When channel investment is minimal, its influence is typically underestimated, while predominant channel investments often yield exaggerated influence measurements—a limitation present in both Bayesian methods and non-hierarchical approaches like Ridge regression \cite{Malthouse1999}.  

\begin{table}[h]
\centering
\caption{Coefficients $a, b$ for Michaelis-Menten-Boltzmann model for dataset 1}
\label{tab:coefficients}
\begin{tabular}{lcc}
\toprule
Media channel & $a$ & $b$ \\
\midrule
Channel 1 offline & 0.941 & 0.0387 \\
Channel 2 offline & 0.940 & 0.0405 \\
Channel 3 online & 0.939 & 0.0562 \\
Channel 4 offline & 0.939 & 0.0561 \\
Channel 5 online & 0.941 & 0.0525 \\
Channel 6 online & 0.945 & 0.0315 \\
Channel 7 online & 0.941 & 0.0447 \\
Channel 8 online & 0.940 & 0.0577 \\
Channel 9 offline & 0.939 & 0.0605 \\
Channel 10 online & 0.940 & 0.0507 \\
\bottomrule
\end{tabular}
\end{table}

Table \ref{tab:coefficients} summarizes the media funnel effects in dataset 1 using the Boltzmann-type equation, where $v_i^*$ represents downstream channels (donors), and $v_i$ upstream channels (receivers). Observing the data, it seems clear that parameter $a$ shows a nearly constant value ($\mu=0.940$ and $\sigma=0.001788$) across all channels, while $b$ shows more variability ($\mu=0.0489$ and $\sigma=0.009594$). It's important to note that these values are strongly dependent on the prior distributions chosen in our model. Drawing an analogy with gas's kinetic theory, a uniform $a$ suggests a consistent modification of individual speeds. This observation implies that all channels share a common property influencing their behavior in a consistent manner. In the context of gases, such uniformity in kinetic parameters could mean analogous molecular properties like mass, molecular size, cross-sectional area, or internal energy, affecting molecular speeds uniformly across the entire system. 

\begin{figure}[H]
\centering
\includegraphics[width=0.5\textwidth]{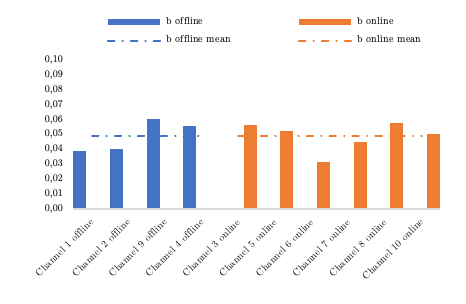}
\caption{Comparison of coefficient $b$ for online and offline channels for dataset 1}
\label{fig:coefficient-b}
\end{figure}
\begin{figure}[H]
\centering
\includegraphics[width=0.65\textwidth]{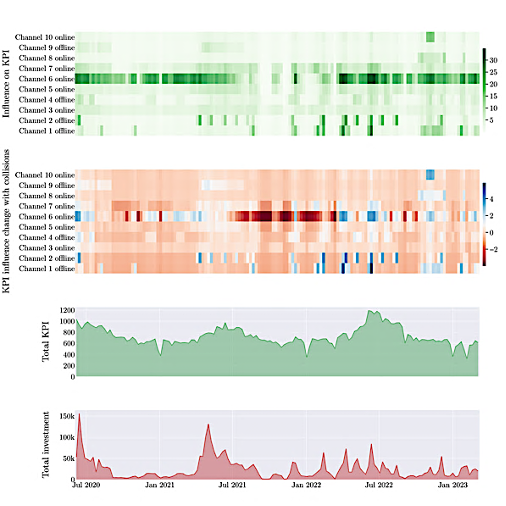}
\caption{Channel collisions impact modeled with Boltzmann equation for dataset 1. Comparison with KPI and total advertising spend}
\label{fig:collisions-kpi}
\end{figure}
\begin{figure}[H]
\centering
\includegraphics[width=0.5\textwidth]{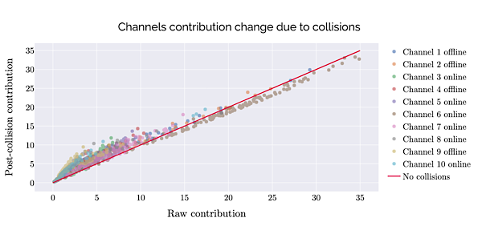}
\caption{Channel collisions impact according Boltzmann equation for dataset 1}
\label{fig:collisions-impact}
\end{figure}


igure \ref{fig:coefficient-b} reveals no statistically significant difference in $b$ coefficient values between online and offline channels. This outcome aligns with expectations, as the $b$ coefficient embeds the combined effect of both online and offline media. Thus, variations in $b$ are unlikely when examined separately for online and offline channels. Figures \ref{fig:collisions-kpi} and \ref{fig:collisions-impact} illustrate the consequences of channel interactions in the context of a $N$-particle system. Notably, Channel 6 online emerges as a "donor" channel, suggesting that its impact extends to other channels. This finding suggests that without this modeling approach, channels may receive incorrect attribution for influence on the dependent variable $y$.

\section{Conclusions}

Our research addresses two critical challenges in Marketing Mix Modeling: investment-biased channel attribution and cross-channel influence quantification. By incorporating the Michaelis-Menten model for shape effects and Boltzmann kinetic theory for funnel effects, we develop a framework that overcomes spending-dependent attribution biases while revealing cross-channel dynamics.
Experimental validation demonstrates that our model maintains comparable predictive accuracy to established MMM approaches while providing superior analytical insights. The normalized Michaelis-Menten constant $\tilde{K_M}$ offers an investment-independent measure of channel effectiveness, eliminating the traditional bias toward high-spend channels. Meanwhile, the Boltzmann-type equation effectively quantifies how each channel influences others' performance—information critical for accurate attribution and strategic resource allocation.

While our integrated model shows promise, implementation within fully Bayesian frameworks introduces challenges derived from increased model parameter dimensionality. Future research should explore simplified formulations that preserve these models' analytical advantages while limiting parameter space complexity. Nevertheless, this physics-inspired approach represents a significant advancement in addressing two persistent challenges in marketing analytics: spend-dependent attribution and cross-channel effect quantification.

\bibliographystyle{plainnat}

\end{document}